\documentclass[preprint,12pt,authoryear]{elsarticle}

\usepackage{graphicx}
\usepackage{url}
\usepackage{apalike}

\journal{BICA}

\begin{document}
\begin{frontmatter}

\title{A Cognitive Architecture for the Implementation of Emotions in Computing Systems}


\author[inst1]{Jordi Vallverd{\'u}}
\ead{jordi.vallverdu@uab.cat}
\author[inst2]{Max Talanov}
\ead{m\_talanov@it.kfu.ru}
\author[inst2,inst3]{Salvatore Distefano}
\ead{s\_distefano@it.kfu.ru, salvatore.distefano@unime.it}
\author[inst4]{Manuel Mazzara}
\ead{m.mazzara@innopolis.ru}
\author[inst2,inst4]{Alexander Tchitchigin}
\ead{a\_tchichigin@it.kfu.ru, a.chichigin@innopolis.ru}
\author[inst4]{Ildar Nurgaliev}
\ead{nurildar9@gmail.com}

\address[inst1]{Universitat Aut{\`o}noma de Barcelona, Catalonia, Spain.}
\address[inst2]{Kazan Federal University, Russia.}
\address[inst3]{University of Messina, Italy.}
\address[inst4]{Innopolis University, Russia.}

\begin{abstract}


In this paper we present a new neurobiologically-inspired affective cognitive architecture: NEUCOGAR (NEUromodulating COGnitive ARchitecture). The objective of NEUCOGAR is the identification of a mapping from the influence of serotonin, dopamine and noradrenaline to the computing processes based on Von Neuman's architecture, in order to implement affective phenomena which can operate on the Turing's machine model. As basis of the modeling we use and extend the L{\"o}vheims Cube of Emotion with parameters of the Von Neumann architecture. Validation is conducted via simulation on a computing system of dopamine neuromodulation and its effects on the Cortex.
In the experimental phase of the project, the increase of computing power and  storage redistribution
due to emotion stimulus modulated by the dopamine system, confirmed the soundness of the model.

\end{abstract}
\end{frontmatter}

\section{Introduction}

In recent years, the complexity and power of the human brain/mind system has been further revelead by several studies, still an uncountable number of questions regarding mechanisms and functions has not been answered. An increasing number of biologically-inspired studies and publications has apperead in the scientific community spanning from approaches to Artificial Intelligence (AI) to results of more general interest in  Information Technology (IT). It has been widely recognized that emotions play an important role in human life \cite{picard1997, fellous2004, marsella2003}. However, traditional AI approaches do not take into account all the dynamics and creative aspects of the mind, therefore demanding for new attacks to the problem of mind modeling.

There are several projects which is worth mentioning in the context of emotions and affect simulation in cognitive architectures: Kansei implemented an emotional-oriented approach \cite{KANSEI}, emotions are widely taken into account in CogAff by \cite{sloman2003}, and work of \cite{OCC} ``OCC'' is based on the appraisal theory and the emotion generation.
The seminal book for whole domain of the ``Affective computing'' was created by \cite{picard1997}, and also located in the MIT we can find one of the first and brightest examples of the social robotics: ``Kismet'' by 
\cite{breazeal2002}.

One of the groundbreaking books on emotions and their implementation in a computing system was written by  
\cite{minsky2007}, where among huge number of ideas he provided the description of the role of  the emotions and identified possible computing implementation approaches.

On the other hand, the neuropsychological picture of emotions in form of base affects was provided by 
\cite{Lovheim2012}, where emotions are expressed in a three-dimensional model of monoamine neurotransmitters: serotonin, dopamine, noradrenaline.

Still a comprehensive framework bridging the biological affects and computing processes is lacking in the scientific literature. In this paper we attempt to fill the gap providing the bio-inspired approach for the mapping of the biological affects on modern computers with common Von Neumann architecture.

The main contribution of this paper is twofold:
i) extending the neuro-psychological three dimensional model of affects, the Cube of Emotion,
with mapping to parameters of von Neumann machine and computing systems;
ii) implementing neuromodulation mechanisms of dopamine in spiking neural network NEST
\footnote{\url{http://www.nest-initiative.org/}} \cite{Gewaltig:NEST} framework, thus one axis of three dimensional model and indicated the connection of one affect (fear) with computing power load and storage parameters.

The remainder of the paper is organized as follows: problem description is provided in  Section \ref{sec:problem}, while our emotion-based approach, NEUCOGAR, is presented in  Section \ref{sec:neucogar}, and detailed analysis is provided in Section \ref{sec:val} with experiment description and results discussion.
Section \ref{sec:concl} closes the paper with some remarks and conclusions.

\section{The problem}
\label{sec:problem}

It has been historically believed that AI systems could have been built upon traditional computing architectures, and that  with enough computing power and memory resources, plus some good algorithms, a human-like symbolic intelligence was feasible. It was just a matter of time, money and technology after all. After the initial excitement to this regard (mainly in the 50s of last century) and numerous failures in reaching the desired results, it has been lately hypothesized that this may
not be the case. Technically, the neuroendocrine system that provides the mechanisms for brain activity may not be a Turing Machine, though the functioning is not fully understood at the moment. In particular, it is not clear whether higher functions performed by humans, as for example consciousness, are epiphenomena of the brain itself or not.

Despite of this, and accepting the fact that we may not be able to model the full neuroendocrine system into the Von Neumann architecture, parallels can still be built, and relationships between the two can help in developing new systems for specific purposes. This is the case of L{\"o}vheim Cube of Emotion where analogies are only built on top of it in relation to \textit{dopamine, serotonin and noradrenaline} brain levels. Simulation of emotions plays a major role in several IT and AI field, not last in the robotics. The ability of a system to feel emotions like, for example, \textit{fear, interest or joy} can trigger behaviors that were not possible in the era of \textit{unemotional machines}. Our interest on emotional AI is not the result of basic curiosity, but is grounded on most of evidences that  show emotions as main regulators of cognitive processes.

The study of brain disorders, related to emotional management, can also provide ways to understand some complex actions like creativity, concentration or interest. Recently, a new generation of cognitive scientists and roboticists proposed a turnover: on one hand, emotions are a fundamental part of the cognitive processes (attention, motivation, strategy selection, mood disposal, reaction, invention, among a long list); on the other hand, the intrinsic relationship between mind and body led to the birth of embodied cognition or grounded cognition \cite{Barsalou2008}. 
This allowed the emergence of a second and powerful wave of cognitive and robotics experts lead by people like 
\cite{Brooks1991, Brooks1995}, 
\cite{Delancey2001} (Robotics), 
\cite{Clark2003} (Philosophy), 
\cite{Damasio1999},  \cite{Ramachandran2004}, and \cite{Rizzolatti2004} (Neuroscience).

In the middle of this cognitive revolution that led to embodied robotics, enactive cognition or morphological computation ideas, an important discipline emerged as a main reference: neuroscience. 
Neuroscientists revolutionized the discipline with data coming from \emph{in vivo} scanned brains (by EEG, fMRI) and brought relevance to the emotional processes into the whole system.  
As a consequence, some authors, such as \cite{Lli}, attempted explanations on the emergence of consciousness. 
Therefore, researchers in AI or cognitive sciences started to introduce ideas about emotional processing.

Once explained the crucial role of grounded and emotional aspects in cognitive architectures, we assume the L{\"o}vheim model as a reference for our own architecture. 
\cite{Lovheim2012} based his theory on a three-dimensional model of emotions and monoamine neurotransmitters (\textit{serotonin, dopamine, noradrenaline}).
The vertexes of the model are affects, as defined by the Tomkins theory, which describes eight basic emotions in \cite{kelly2009}.
Tomkins labeled with one word for the emotion when it was of low intensity and another word for the same emotion at a higher intensity.
\cite{tomkins1962,tomkins1963,tomkins1991} referred to basic emotions as ``innate affects'', where affect, in his theory, stands for the ``strictly biological portion of emotion".
According to this theory,  the eight basic emotions are: \textit{enjoyment/joy, interest/excitement, surprise, anger/rage, disgust, distress/anguish, fear/terror and shame/humiliation}.
Some of them are managed only by the internal processes of the individual, the others are intrinsically related to social interactions.

\section{Our cognitive affective architecture: NEUCOGAR}
\label{sec:neucogar}

In this section we present our novel approach to artificial cognitive architectures, the NEUromodulating COGnitive ARchitecture (NEUCOGAR). The idea is to create a mapping of monoamine neuromodulators to emotional states
applied to computing system parameters. Our work starts from the following considerations:

\begin{enumerate}
\item Emotions are natural and necessary modulators, reinforced and modeled at the same time by social external
factors or even internal thoughts.
\item Monoamine neuromodulators are the physical mechanism by which emotional responses
are triggered and modulated within cognitive systems like humans, \cite{marsella2003, marsella2010}.
\item
Several modular models are available: Geneva emotion wheel, Plutchick's ``Wheel of emotions'' \cite{plutchik2001},
or  Cube of Emotion \cite{Lovheim2012}, among others. 
All of them identify different strategies, the basic emotions and how dynamic transitions among them are created \cite{gratch2005}.
\end{enumerate}

Stemming from this simple points, our work is focused on modeling the mapping of impacts of monoamine
neuromodulators on human brain into the computing processes of modern computers.
We propose to correlate biochemical influence of monoamines such as \textit{dopamine, serotonin,
and noradrenaline}, involved in affective processing of cortical, limbic and other subsystems of human brain with computing system parameters related to \textit{computing power, memory distribution, storage}.

\subsection{From real emotions to artificial ones}

The idea behind the mapping of neuromodulators to computing system parameters stands
in creating a link between the Von Neumann architecture (abstractly a Turing Machine) and the neuroendocrine system. 
Our research makes possible to implement 
a biologically inspired architecture based on neuromodulators. 
As we will show during the next sections, this innovative approach can provide new ways to automatically manage and devise 
very important features performed by human beings but still far from computing feasibility. 
With our research we are not trying to propose a new version of a cognitive architecture  that attaches an emotional module \cite{emocog} or \cite{kismet}. 
Our aim is radically different, \cite{vallverdu}: building an emotional structure that is at the same time the cognitive structure.
 As just in human beings and most of living entities.

As the human brain as a whole reacts to changes in the neuromodulator and hormonal levels, the model
presented in this paper offers a ground for simulations of emotions in the Von Neumann architecture
(with the clear understanding that, most likely, not every brain function can be reproduced in the same
way). This mapping is clearly not sufficient: proper algorithms need to be developed to reproduce higher
functions on top of the \textit{``hormonal reactive machinery''}. These algorithms will monitor CPU activity
and memory allocation in order to track (simulated) \textit{hormonal and neuromodulators levels} and react accordingly.

Higher functions of the brain like consciousness and proactivity (the ability of not reacting in the short term
to immediate physiological responses) may possibly be simulated on top of basic functions. However, this
still remains an unproven hypothesis needing accurate investigation. If such higher functions are indeed
epiphenomena of the brain (or endocrine systems) is still under debate in several scientific communities.

According to their nature  we can split computing system parameters in three groups:
processing, memory and storage.
More specifically, the parameters taken into account are:

\begin{description}
 \item[Computing utilization] is a metric able to quantify how busy the processing resources of the system are. It can be expressed by the average value of all the single processing resources' utilization.
 \item[Computing distribution] aims at quantifying the load balancing among processing resources. It can be expressed as the variance of single resources' utilization.
 \item[Memory distribution] is associated with the amount of memory allocated to the processing resources. It can be quantified by the variance of the amount of memory per single resource.
 \item[Storage volume] is an index related to the the amount of data and information used by the system.
 \item[Storage bandwidth] quantifies the number of connections between resources, i.e. processing and data nodes.
\end{description}

\subsection{NEUCOGAR in details}
\label{sec:analysis}

\begin{figure}
    \centering
    \includegraphics[width=0.8\columnwidth]{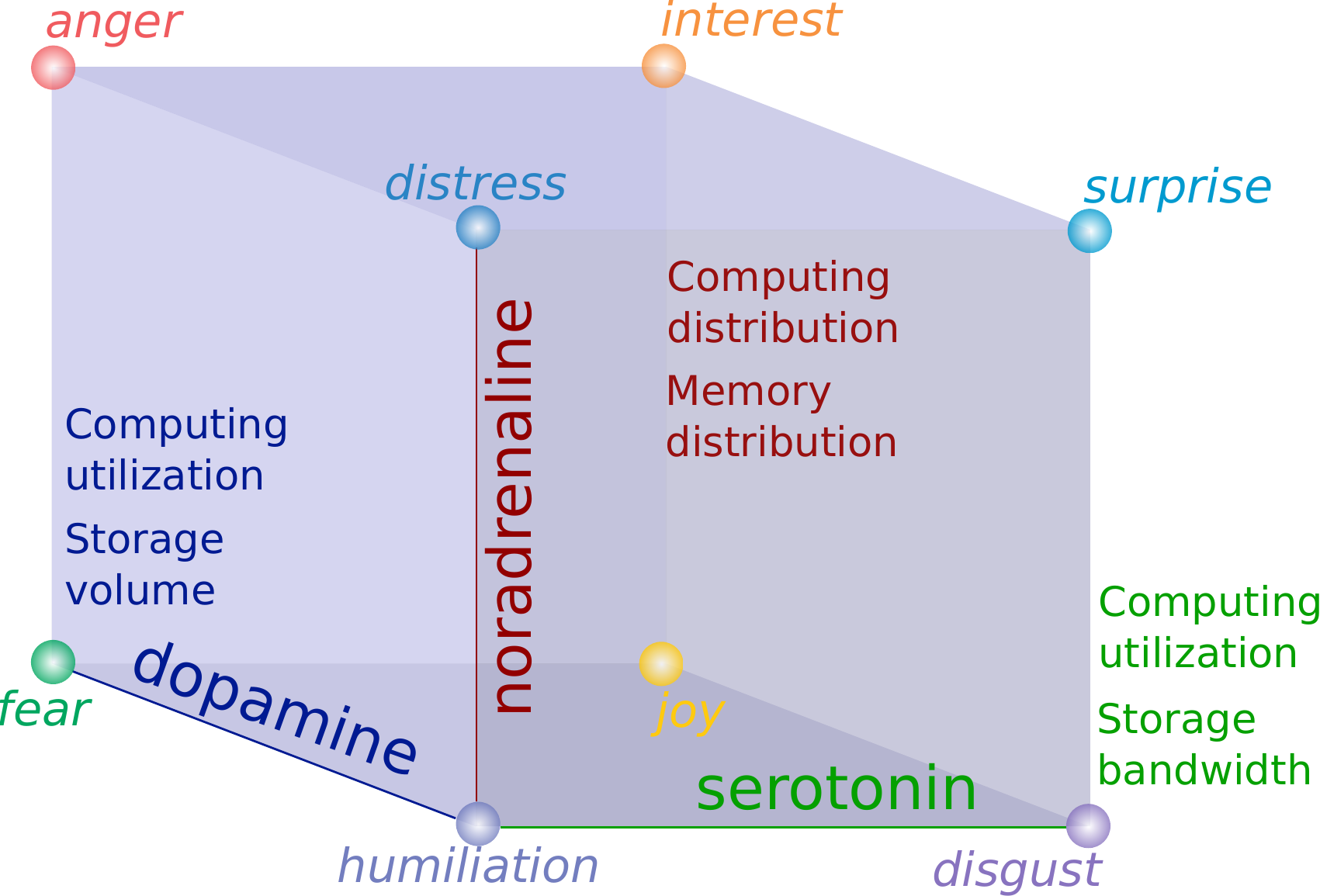}
    \caption{The mapping of the \cite{Lovheim2012} Cube of Emotions  to computing system parameters}
    \label{fig:cube}
\end{figure}

In Fig. \ref{fig:cube} a revised version of the Cube of Emotion model is depicted as originally introduced in \cite{Lovheim2012}.
 The information on the cube is adapted to the computing domain and extended to include  parameters related to computing power, memory and storage.
It represents the link bridging neuroscience and psychology with concepts of computing and AI systems.
This paper aims at building a common understanding between these three worlds and possibly fostering a productive cooperation among specialists of all these fields.

Low level influence of affects on brain (cellular or neuronal) provides fundamental base for our affective
computation model. We draw an analogy between computing processes in computer and neurotransmission in brain, simulating
biochemical influence of neuromodulatory systems on neurons and translating them into the computing processes
of a machine.
Especially interesting from our perspective is \cite{fellous2004}, where the role of dopamine,
serotonin and their impact in emotional context (of mammalian brains) is discussed.
It can be considered as one of the basis for our work in the simulation of neuromodulation domain.
Our interest for these 3 neurotransmitters is based on the fact that they are the three main chemical substances managing brain's inhibitory and excitation processes that drive and run cognition, attention, and perception.
Furthermore, serotonin and noradrenaline strongly influence mental behavior patterns, while dopamine is involved in movement.
For this reason they are the focus of neuroscientific studies for many years  \cite{neurotransmitters}.

More specifically:
\emph{Dopamine} appears to play a major role in motor activation, appetitive motivation, reward processing and neuronal plasticity, and therefore might be important in emotion. 
According to L{\"o}vheim Cube of Emotion, dopamine is associated with \emph{``reward, reinforcement, motivation''}. 
In the computational interpretation 
presented in this work, dopamine plays a rewarding role in computing utilization and storage volume, i.e. increased overall processor activity and amount of memorized data.

``\emph{Serotonin}'' - L{\"o}vheim says - \emph{has been implicated in behavioral state regulation and arousal, motor pattern generation, sleep, learning and plasticity, food intake, mood and social behavior''} \cite{Lovheim2012}. 
Serotonin plays a crucial role in the modulation of aggression and in agonistic social interactions in many animals. L{\"o}vheim associates serotonin with \emph{``self, confidence, inner strength, and satisfaction''}.

The computational interpretation of the L{\"o}vheim Cube of Emotion places serotonin as influencer of computing utilization and storage bandwith, i.e. increased overall processor activity (sinergically with dopamine) and rise of connectivity between nodes.


Finally, L{\"o}vheim Cube of Emotion emphasizes the \emph{noradrenaline} role in: \emph{``Attention,
vigilance, activity''}.  Then, it describes the role of noradrenaline: \emph{``... noradrenaline has been coupled to the fight or flight response and to stress and anxiety, and appears to represent an axis of activation, vigilance and attention''}.
In the computational interpretation, noradrenaline mainly influences computing and memory distribution, since it allows to promptly react to some specific conditions and/or alerts, changing the configuration of the system at runtime, and therefore impacting on processing and memory configuration.

A holistic view of the diagram shows that:
\begin{description}
 \item[Computing utilization] is influenced by dopamine and serotonin, which increase the firing spike activity of cortex neurons and therefore increase usage of the ``artificial'' brain of the computing system.
 \item[Computing distribution] is influenced by noradrenaline. The higher the level of noradrenaline, the more computing power has to be concentrated on the current activity (neuromodulator regulating attention).
 \item[Memory distribution] (short term memory/focus) is impacted by noradrenaline, which is moudulator of attention.
 \item[Storage volume]  (long term memory) is impacted by both serotonin and dopamine, i.e. higher concentrations of
both neuromodulators makes system more efficient in remembering a stimulus. In general, strong emotions generate more persistent memories and involve bigger amound of data.
\item[Storage bandwith] is influenced by serotonin, i.e. higher concentrations of the neuromodulator increases the connectivity, i.e. the number of connections between nodes.
\end{description}

\section{Validation}
\label{sec:val}
To the purpose of validating our approach, we have implemented the dopamine neuromodulation of the fear affect in the computing system and check one axis of our model.
In this section we report on preliminary experiments we performed.
Specifically, in Section \ref{sec:experiment} we describe the setup of the experiment, while in Section \ref{sec:results}
we present the obtained results.

\subsection{Experiment description}
\label{sec:experiment}

\begin{figure}
\center\includegraphics[width=0.6\textwidth]{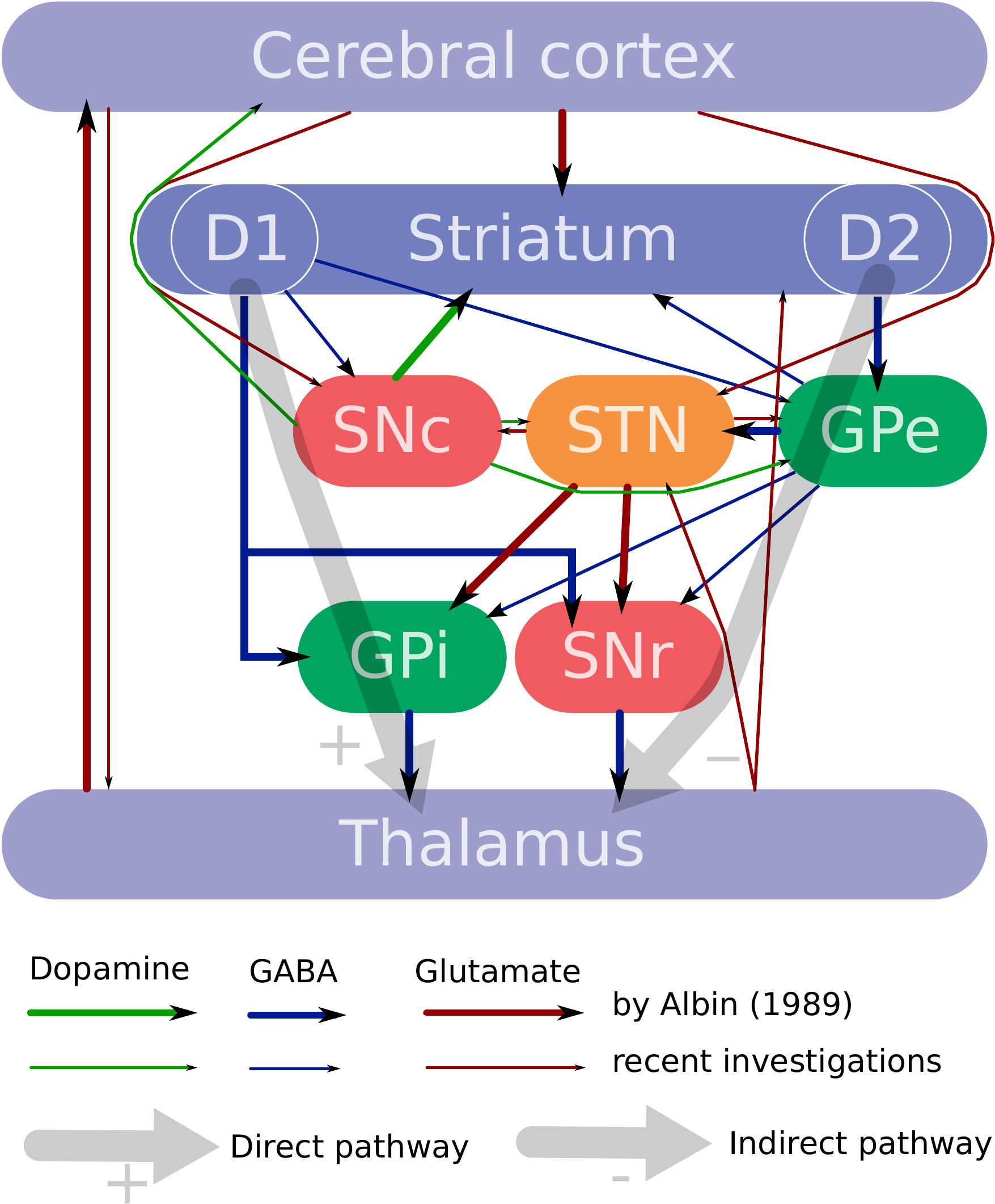}
\caption{Nigrostriatal dopamine pathway: GPe\ --- Globus pallidus external; GPi\ --- Globus pallidus internal; STN\ --- Subthalamic nucleus;
SNc\ --- Substantia nigra compacta; SNr\ --- Substantia nigra reticulata.
Red arrows identify excitation (glutamate), blue --- inhibition (GABA), green --- modulation (dopamine)
connections. Striatum balances towards direct pathway (gray arrow with +) due to influence of the dopamine from the SNC (green arrow).}
\label{fig:BG_advanced}
\end{figure}

The experiment looks for the validation of dopamine neuromodulation and its influence on a cerebral cortex and therefore thinking processes.
We made the general assumption that thinking processes are analogous to or could be expressed as computing processes.
We have implemented a \emph{nigrostriatal dopamine pathway} published by \cite{dopa_albin} with some extensions for feed-back loops and additional connections according to modern findings in neuroscience as shown in Fig. \ref{fig:BG_advanced}.
The dopamine neuromodulation is modulated by an area of the brain called \emph{substantia nigra compacta} (SNc) that produces dopamine and projects it to another brain area named striatum for balancing between pathways.
The \emph{mammalian striatum} consists of neurons with different dopamine receptors: D1 and D2 
balancing
direct and indirect pathways (depicted below in Fig. \ref{fig:BG_advanced}) through  their heterogeneous sensitivity to dopamine neurotransmitter.

The direct pathway is
\textit{Cerebral cortex} (stimulate) $\to$ \textit{Striatum} (inhibit) $\to$ \textit{complex SNr-GPi} (Thalamus is less inhibited) $\to$ \textit{Thalamus} (stimulate) $\to$ \textit{Cerebral cortex} (stimulate) $\to$ \textit{muscles and etc.}

The indirect pathway is
\textit{Cerebral cortex} (stimulate) $\to$ \textit{Striatum} (inhibit) $\to$ \textit{GPe} (STN is less inhibited) $\to$ \textit{STN} (stimulate) $\to$ \textit{complex SNr-GPi} (inhibit) $\to$ \textit{Thalamus} (is less stimulated) $\to$ \textit{Cerebral cortex} (is less stimulated) $\to$ \textit{muscles and etc.}\\

\subsection{Results}
\label{sec:results}
\begin{figure}[ht]
\centering
  \centering
  \includegraphics[width=0.9\textwidth]{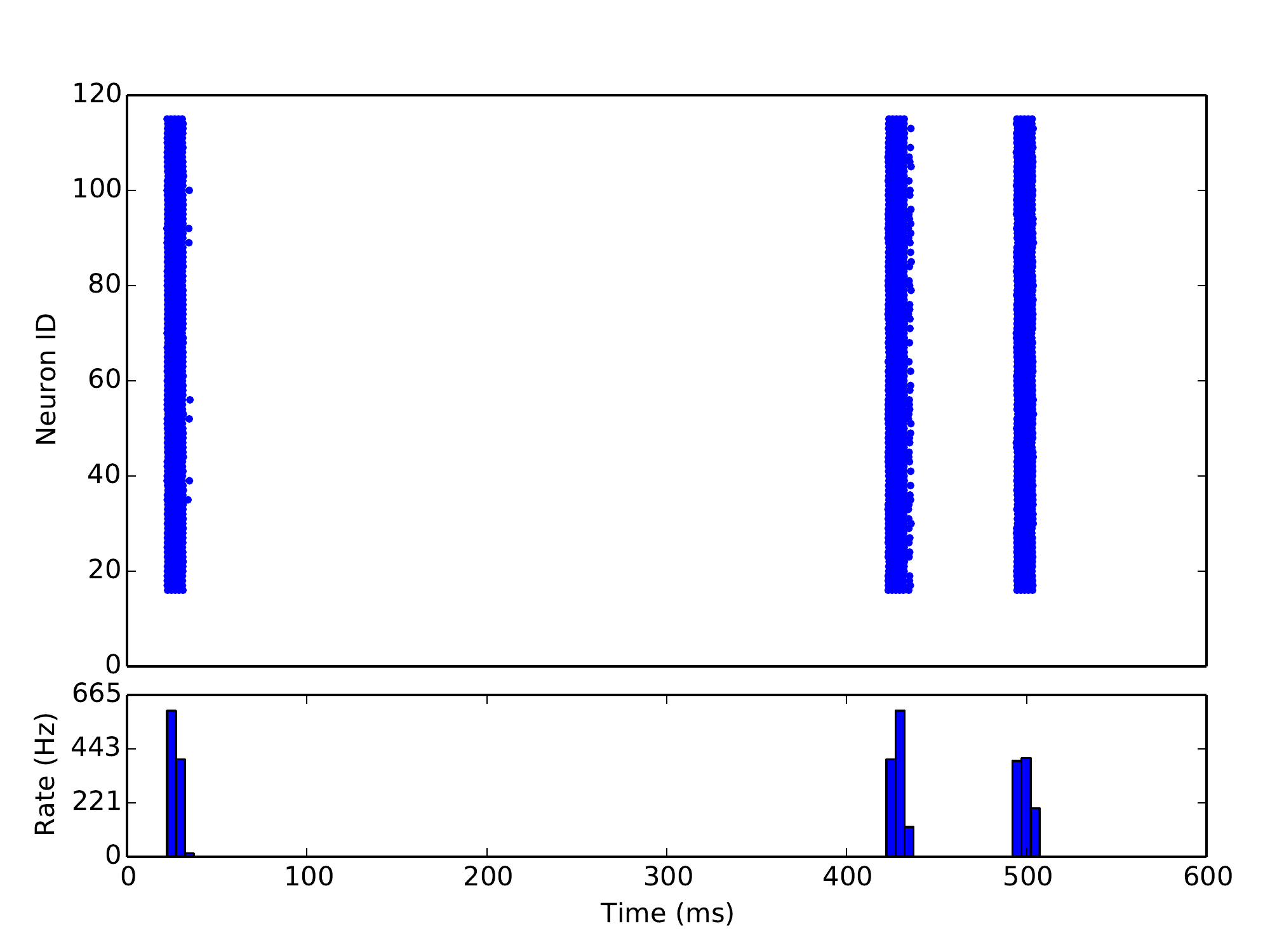}
  \caption{The graph of spiking activity on the simulated Thalamus from the nigrostriatal pathway. The increased activity is seen from 400 to 550 ms, while the dopamine neuromodulation was initiated at 400 ms.}
\label{fig:thalamus_activity}
\end{figure}
\begin{figure}[ht]
  \centering
  \includegraphics[width=0.9\textwidth]{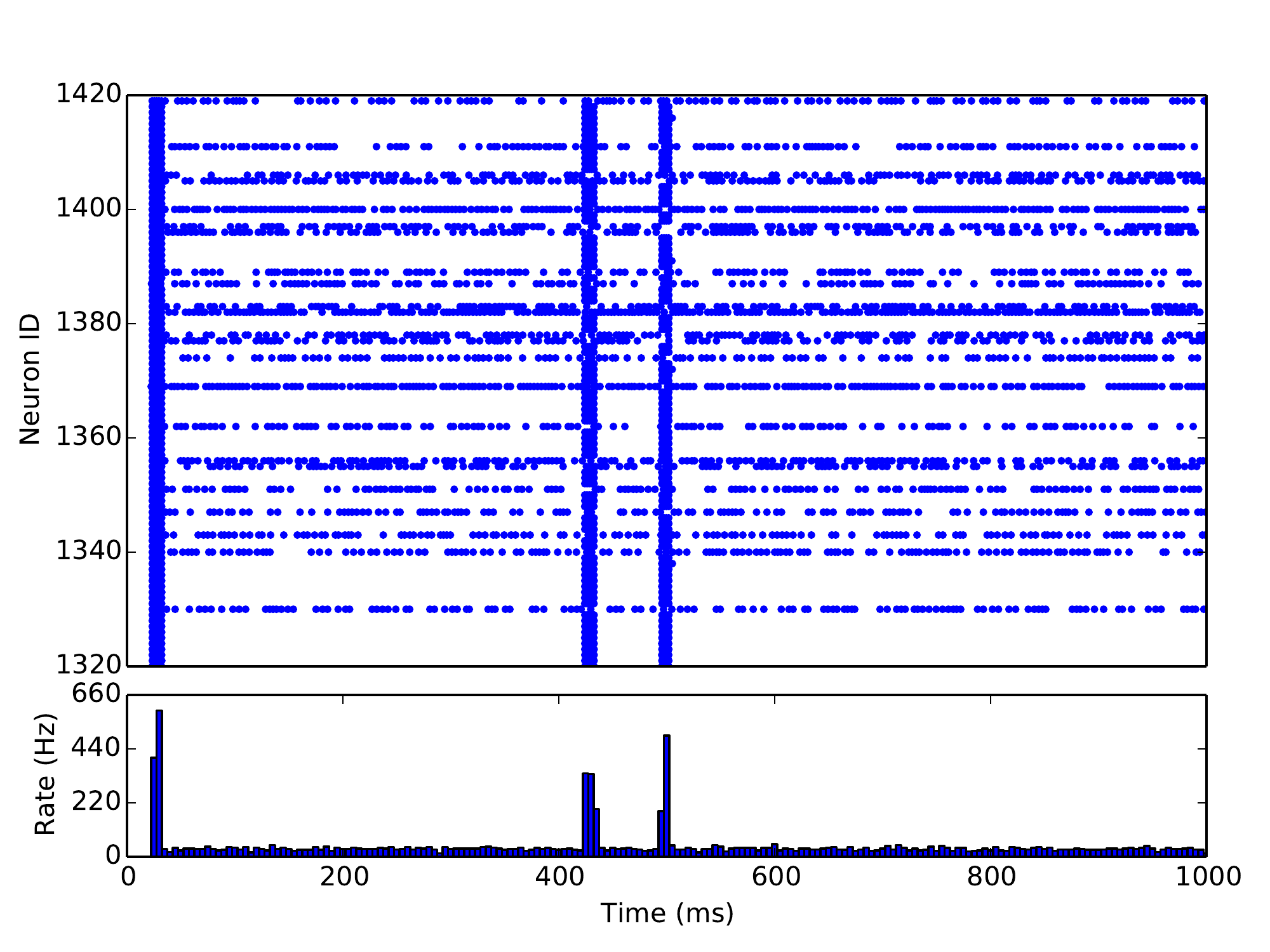}
  \caption{The graph of spiking activity on the simulated motor cortex from the nigrostriatal pathway. The increased activity is seen from 400 to 550 ms, while the dopamine neuromodulation was initiated at 400 ms.}
\label{fig:cortex_activity}
\end{figure}
In our simulation through the spiking neural network tool NEST, we aimed at demonstrating these mechanisms, with specific focus on dopamine neuromodulation.
In particular, during the experiments, we observed the spiking activities of thalamus and cerebral motor cortex reacting to these neromodulation.
We therefore created a spiking network model of the overall nigrostriatal pathway simulating its behaviour in the [0,1000]ms simulation time frame.
At 400ms we triggered a volume transmission of dopamine in substancia nigra and observed the effects in both thalamus and cerebral motor cortex.

Fig. \ref{fig:thalamus_activity} describes the reaction of the thalamus.
It shows the increasing spiking activities right after the input from substancia nigra, highlighting a growing activity on computing power, increasing its utilization.
As a side effect, this enhanced the synapses among neurons thus enforcing and increasing data persistence in storage.

With regard to the cerebral motor cortex, Fig. \ref{fig:cortex_activity} reports its activity.
As for thalamus, from this plot we can easily identify a reaction to the input at time 400ms due to the interaction with thalamus.
The raster plot also shows the noise activity typical of mammalian cerebral cortex.

The above results concur to demonstrate the correctness of our approach based on the hypotheses and on the model discussed in  Section \ref{sec:neucogar} along to only one axis   of the Cube of Emotion associated with dopamine.

\section{Conclusions and future work}
\label{sec:concl}

The mind/body management is related to emotional signals that capture, process and classify data inputs and allow responses.
We propose the neurobiologically inspired cognitive architecture for the simulation of  affects in a computing system.
Our approach is innovative for several reasons:

\begin{itemize}
\item We need not running a whole brain simulation in order to achieve a functional
artificial cognitive system, and this allow us to avoid the still existing and deep
uncertainties between brain structure and performance
(at this historical moment, brain researchers are at the same point that genetic
biologists starting to decode the genome: with a lot of information, in fact too much,
but without meaning).
\item We are not qualitatively imitating the surface of the emotional mechanism of
human brains and bodies, instead we are suggesting a new approach that is
rooted in the neuromodulation, the key piece of brain emotional and
cognitive processes. It is out of any doubt that basic as well as high cognitive performance skills are related to emotional management, which makes possible innovative, creative and fresh ways to deal with multihaptic and raw data. Emotional architectutre for AI is not the result of interests on HRo or some human psychological processes, but the real necessity of creating better autonomous systems. Data mining and deep learning are providing new connections between data, but not new wways to deal with data, something that we expect that emotional architectures will be able to perform.
\item This functional approach allows generation of more dynamic systems
that add cognitive power beyond the basics of classic weights, thresholds or
valences, providing options to build a cognitive architecture that generates several faces
of emotional interactions such as moods, attitudes, emotional memories or affective states.
\item This approach makes possible to implement computingly a more subtle training mechanism of artificial cognitive architectures. Here, the syntax is dynamic but at the same time hardwired into the basic objects. Artificial neurotrasmitters not only flavour emotionally all objects present into the program, but also makes possible more complex interactions. Thus, it open the door towards the integration of multiheuristic approaches under the same architecture, in the same way that operate human beings (using several formal heuristics, abductive, deductive, inductive thinking,...): we will be able to implement a system that follows several reasoning rules but that at that time, these rules will be affected by the distribution, concentration and influence of synthetic neurotransmitters.
\end{itemize}

The NEUCOGAR provides the proper background to the implementation of emotional elements into the whole system,
acting as the basic piece of regulation, and not as an attached module with emotional
contents. This architecture allows to apply it smoothly from scratch (the artificial
neuron or unit of processing) to the superior layers of specialized cognitive processes
(like memories, processors). In this sense, NEUCOGAR is a whole artificial cognitive
system that simulates the basic running of bodily emotional systems, following
a modular architecture that is put together and ruled by the same
emotional cyber-neuromodulators.

The validation of the NEUCOGAR is complex task that involves use of spiking neural network
NEST\footnote{\url{http://www.nest-initiative.org/}} \cite{Gewaltig:NEST} and creation
of proper brain cortical and subcortical areas and connections.
We have managed to validate only one axis of the NEUCOGAR model, the dopamine axis, because it has been better explored than others at the moment.
This validation indicated principal correctness of the approach to simulate emotions in the form of inborn affects in a computing system.
In the experiments it has been noted the rise of computing power and of storage redistribution
due to emotion stimulus modulated by dopamine system.
This demonstrated the soundness of the proposed approach on one axis of the Cube of Emotion model.

We consider adding serotonin pathways as the most important extension of current implementation.
That will allow validation of up to 4 affects: joy, fear, disgust, humiliation,
and overall will make the model from one-dimensional to bi-dimensional.
Further enhancement with noradrenaline will provide the 8 basic affects that
could be used as the base for the implementation of the infant emotional cognitive
architecture as well as be helpful for the emotion recognition, evaluation and measurement.

\textbf{Summary - }
We can summarize NEUCOGAR achievements as follows:
\begin{itemize}
\item It provides the first computational bottom up or grounded emotional architecture, which integrates fundamental emotional mechanisms into artificial cognitive systems. In this model, hormonal mechanisms permeates the totality of the system, instead of being one of the modules under interaction, as usual.
\item It is a bio-inspired architecture, based on neurotransmitters, that makes possible to implement human-like cognitive processes into Von Neumann's architectures.
\item The neurotrasmitters provide the system with a unique, flexible,  valuable and  tunable affective model.
\item The validation of NEUCOGAR 
showed its reliability and efficiency at modular level.
\item NEUCOGAR helps us to understand at a scalable level the mechanisms of proactivity, attention, processing, concentration, storage and reaction, among other cognitive processes. 
It is the functional and best reverse approach for the naturalization and functional improvement of AI.
\item NEUCOGAR  allows to start to think about the mechanisms that act during the switch between multiheuristic strategies within cognitive systems (as individuals as well as a social collectives).
\end{itemize}

\section*{Acknowledgments}

Partial support for this research was received by the Spanish Government's DGICYT research project: FFI2011-23238, ``Innovation in scientific practice: cognitive approaches and their philosophical consequences" and by the Russian Ministry of education and science (agreement: 14.612.21.0001, ID: RFMEFI61214X0001).

\bibliographystyle{apalike}
\bibliography{neucogar}

\end{document}